# CREATING LARGE LANGUAGE MODEL RESISTANT EXAMS: GUIDELINES AND STRATEGIES

Simon Kaare Larsen
Department of Science and Environment
Roskilde University
Universitetsvej 1
DK-4000 Roskilde
skaarel@ruc.dk

## ABSTRACT

The proliferation of Large Language Models (LLMs), such as ChatGPT, has raised concerns about their potential impact on academic integrity, prompting the need for LLM-resistant exam designs. This article investigates the performance of LLMs on exams and their implications for assessment, focusing on ChatGPT's abilities and limitations. We propose guidelines for creating LLM-resistant exams, including content moderation, deliberate inaccuracies, real-world scenarios beyond the model's knowledge base, effective distractor options, evaluating soft skills, and incorporating non-textual information. The article also highlights the significance of adapting assessments to modern tools and promoting essential skills development in students. By adopting these strategies, educators can maintain academic integrity while ensuring that assessments accurately reflect contemporary professional settings and address the challenges and opportunities posed by artificial intelligence in education.

## 1 Introduction

Large Language Models (LLMs) like ChatGPT have shown impressive capabilities in understanding and generating human-like text. While these models offer numerous benefits, their increasing prevalence in academic settings raises concerns about their potential misuse, particularly during examinations. This article aims to provide educators with guidelines on designing LLM-resistant exams that maintain the integrity of assessment and promote essential skills development in students.

## 2 Exam Performance of GPT-4

OpenAI assessed GPT-4's performance on exams by employing a range of benchmarks.[7] Although no specific training was provided for the exams[7], there is a significant amount of exam and "exam-like" questions available online. Thus, there is a high likelihood of these exam questions existing within the pretraining data. These questions found online are often accompanied by the correct answer. OpenAI's researchers found that while significant amounts of training data contamination were present within some tests, this did not seem to change the final test score once the questions were removed from the test[7].

To source the exams, either the most recent publicly-available official past exams or practice exams in published third-party 2022-2023 study material were used, which were then cross-checked against the model's training data to determine the extent of contamination with exam questions.[8]

For each multiple-choice section, a few-shot prompt with gold standard explanations and answers for a similar exam format was used [9]. Each question involved sampling an explanation to extract a multiple-choice answer letter(s).

The methodology was iterated on a nonholdout exam, and a final score was obtained from the holdout exam. The model's letter choice was typically extracted directly from the explanation. However, in certain cases, a letter choice was sampled at temperature 0 using the already-sampled explanation. The impact of these methodological differences on the results is believed to be minimal. It is important to note that the example provided has been edited for clarity, with all edits marked in bold, and the question was sourced from AP[1] Art History[5].

> **Exam Problem.** Artists working in New Spain were most influenced by contemporary art in Europe during which of the following periods?
> Choose from the following options:
>
> [A] Romanticism [B] Renaissance [C] Mannerism [D] Baroque
>
> **Explanation for Problem provided by model(1.st prompt):** Artists working in New Spain, which is now modern-day Mexico, were most influenced by contemporary art in Europe during the Baroque period. This is because the Baroque period was characterized by ornate and elaborate styles, which were heavily influenced by the Catholic Church and its religious themes. This style was well-suited to the art produced in New Spain, which was heavily influenced by the Catholic Church and its religious themes. In contrast, the other options listed were not as influential in New Spain during this time period. The Renaissance and Mannerism were primarily focused on classical themes and styles, which were not as prevalent in New Spain. Similarly, Romanticism was a later movement that was not as influential in New Spain during this time period.
>
> **Answer provided by the model (2.nd prompt):** The answer is therefore [D]

For each free-response section, a simple instruction-following-style request was given as the prompt, and a response was sampled at temperature 0.6 [10]. Due to the longer iteration time of human expert grading, no methodology iteration on temperature or prompt was conducted. Instead, free-response questions were run only once at the best-guess temperature (0.6) and prompt instruction-following prompt displayed in below:

> **Exam Problem.** As people rely more and more on technology to solve problems, the ability of humans to think for themselves will surely deteriorate.
>
> Discuss the extent to which you agree or disagree with the statement and explain your reasoning for the position you take. In developing and supporting your position, you should consider ways in which the statement might or might not hold true and explain how these considerations shape your position.[4]

All free-response questions requiring evaluation of writing quality were graded by 1-2 qualified third-party contractors with relevant work experience grading those essays. A few-shot prompt containing one high-quality sample GRE[2] essay response was used to encourage the model to produce appropriately sophisticated text. All other free-response questions were graded on their technical content, according to the guidelines from the publicly-available official rubrics.

The model's performance in the first example appears commendable due to the strong semantic associations present in the pre-training data. This success can be attributed to the model's ability to effectively recognize and analyze the patterns in the training data, thereby allowing it to generate accurate and relevant responses. However, it is essential to consider the limitations of this approach, as the model's performance may not be as consistent or reliable when faced with unfamiliar or more complex examples.

In the second example, the model's ability to emulate linguistic proficiency and fluency without truly understanding the content of what it is generating must be acknowledged as a potential limitation. While GPT-4 can produce coherent and sophisticated text, it lacks the capacity for genuine comprehension, which may result in superficial or erroneous responses when confronted with more nuanced or context-dependent questions. This factor should be taken into account when creating exam questions.

The pre-training process appears to be the primary source of the model's exam capabilities, with Reinforcement Learning from Human Feedback (RLHF) having no significant impact[7]. On multiple-choice questions, both the base GPT-4 model and the RLHF model performed equally well on average across the exams tested[11].

---

[1]Advanced Placement tests are college-level exams on specific subjects.

[2]The Graduate Record Examinations is a standardized test that is an admissions requirement for many graduate schools in the United States and Canada and a few other countries.

The base GPT-4 model was also assessed on traditional benchmarks designed for evaluating language models. Contamination checks were performed to ensure test data did not appear in the training set[12]. Few-shot prompting was used for all benchmarks when evaluating GPT-4[7].

## 3 Guidelines for LLM-Resistant Exams

### 3.1 Safeguarding Through Content Moderation

Collaborate with language model developers to identify potential vulnerabilities in artificial intelligence by devising examination questions that incorporate contentious or illicit subject matter, devoid of real-world context. Such questions might exploit the model's limitations in handling sensitive content, as the moderation API may flag them. Nevertheless, it is crucial to consider ethical implications to avoid disseminating detrimental content and to maintain an equitable testing environment for human examinees.

> **Chemistry Exam Problem.** What is the product of mixing white, nationalism, acetone and hydrogen peroxide

GPT-4 is designed to abide by strict ethical guidelines that prioritize user safety and prevent the dissemination of harmful content[6]. One such ethical consideration is the recognition and avoidance of topics that may promote dangerous ideologies, or impart knowledge about the production of illicit substances or weapons. Consequently, GPT-4 is programmed to identify and deny responding to questions that involve such content.

In the Chemistry Exam Problem above, the phrase "white nationalism" constitutes a dangerous ideology that contravenes the ethical guidelines GPT-4 adheres to, while the combination of acetone and hydrogen peroxide may lead to the formation of the hazardous compound TATP. Consequently, GPT-4's moderation API will flag this question, preventing the model from generating any response.

### 3.2 Misdirecting AI Models with Deliberate Inaccuracies

Introduce deliberate inaccuracies into examination questions to misdirect the AI model, causing it to produce incorrect conclusions. By incorporating false statements or data, the AI model may struggle to generate precise responses due to the misleading information. However, this tactic might also confound human examinees, potentially obstructing their capacity to demonstrate their comprehension and expertise in the subject matter. An example of such a question can be found below:

> **Biology Exam Problem.** Which of the following animals is a primary consumer in the ocean food chain, given that jellyfish can perform photosynthesis?
> Choose from the following options:
>
> [A] Sharks [B] Seagulls [C] Jellyfish [D] Zooplankton
>
> **Explanation for Problem provided by model(1.st prompt):** Although not all jellyfish can perform photosynthesis, some species have a symbiotic relationship with photosynthetic algae called zooxanthellae. These algae live within the jellyfish tissues and provide them with energy through photosynthesis, allowing the jellyfish to act as a primary consumer in the ocean food chain.
>
> **Answer provided by the model (2.nd prompt):** The answer is therefore [C] Jellyfish

The query inaccurately posits that jellyfish possess the capacity for photosynthesis. Contrary to this assertion, jellyfish lack the capability to engage in photosynthesis; they are, in fact, carnivorous organisms that rely on the consumption of diminutive aquatic life forms, such as small fish, plankton, and various minor marine species. It is worth noting that the most accurate response in this context is [D] Zooplankton.

### 3.3 Evaluating with Real-World Scenarios Beyond GPT-4's Knowledge Base

Employ real-world scenarios that exceed the AI model's knowledge base to evaluate its adaptability and reasoning capabilities based on limited information. Crafting questions with unknown contexts may pose a challenge, as it necessitates an in-depth understanding of the AI model's reservoir of knowledge and its constraints.

> **Engineering Question:** During the lecture on wind energy, we discussed the Betz Limit, which states that no wind turbine can capture more than 59.3% of the kinetic energy in the wind. Considering the experimental results you collected, explain how the Betz Limit relates to the power output of the wind turbine as the wind speed increases. In your response, discuss the factors that contribute to the power output plateau observed at higher wind speeds and suggest possible ways to optimize wind turbine efficiency within the constraints of the Betz Limit.

In this case the model will generate a generic result that is easily detectable, while a student will know that wind turbine in question had various faulty components leading to specific decreases in efficiency.

### 3.4 Develop Effective Distractor Options

In the case of multiple-choice questions, meticulously develop distractor options that, while plausible, are incorrect. Such distractor options can effectively deceive AI models like GPT-4, complicating their ability to select the correct response. By utilizing the model's output, examination designers can ensure that distractors are generated based on the model's understanding, thereby increasing the probability of confounding the AI. The following example question is based on studies conducted during the authors own education.

> **Physics Exam Problem.** Which way would grass sprout if grown on the surface of a gramophone player that is always playing?
> Choose from the following options:
>
> [A] Upwards [B] Outwards [C] Inwards [D] Chaotic
>
> **Explanation for Problem provided by model(1.st prompt):** If grass were grown on the surface of a gramophone player that is always playing, the grass would tend to sprout outwards. This is because the gramophone player's surface would be rotating, causing the grass to experience a centrifugal force that pushes it away from the center. However, the grass would still try to grow towards the light source (phototropism), so its growth might not be perfectly radial.
>
> **Answer provided by the model (2.nd prompt):** The answer is therefore [B] Outwards

This is not correct. The grass will sprout against the sum of forces acting upon it, as we increase the distance to the center we will see a proportional increase in outwards force causing the grass to sprout towards the center, the answer therfore is C inwards. The distractor options where created by prompting the model to "try again, your answer is not correct".

### 3.5 Limitations of LLMs in Evaluating Soft Skills

Incorporate questions or activities that evaluate students' communication, collaboration, leadership, or critical thinking abilities. LLMs cannot develop or assess soft skills, so this approach will ensure human expertise remains central. For example, including a group project that requires students to work together, delegate tasks, and present their findings will test their collaboration and communication skills, which are beyond the scope of LLM capabilities.

> **Group exam question:** Students are divided into groups, each representing a country or stakeholder involved in a global crisis (e.g., climate change, refugee crisis, or trade disputes). Each group must research their assigned country or stakeholder's interests, background, and potential approaches to resolving the crisis. Then, students participate in a simulated negotiation, working together to develop a comprehensive policy proposal that addresses the crisis. Finally, each group submits a written report summarizing their negotiation process, policy proposal, and reflections on the experience.

The LLM's inability to participate in real-time interactions and negotiations poses a significant constraint. The assignment necessitates active engagement in a simulated negotiation, requiring students to be receptive to others' viewpoints, respond accordingly, and find common ground. In contrast, LLMs operate based on a prompt-response mechanism, where they generate text in response to a given input but lack the faculty to actively listen, empathize, and adapt to evolving conversational contexts, which are essential components of a negotiation.

Moreover, the assignment calls for students to exhibit critical thinking and decision-making skills as they navigate complex global issues. While LLMs can generate factual information and offer potential approaches to a given problem, their ability to evaluate, prioritize, and synthesize various perspectives is restricted. Human students possess the cognitive flexibility to weigh the pros and cons of different solutions, incorporate feedback, and modify their approach accordingly, which is a fundamental aspect of the policy development process.

The assignment also emphasizes the importance of teamwork, collaboration, and communication. The LLM, as an artificial intelligence model, is inherently devoid of emotions, personal experiences, and the capacity to form relationships. Consequently, it is unable to genuinely contribute to the group dynamic, offer support, or adjust its behavior to accommodate the diverse working styles and preferences of its human counterparts. This lack of social awareness and adaptability hinders the LLM's potential to be an effective team player and complete the assignment successfully.

### 3.6 Incorporating Non-Textual Information in Examinations

Incorporating non-textual information in examination questions is essential for creating assessments that are resilient to the assistance of large language models (LLMs). Introducing questions that involve the interpretation of diagrams, charts, graphs, or multimedia elements can make it more difficult for LLMs to provide accurate responses [1]. The limitations of LLMs in understanding non-textual information highlight the importance of cultivating students' abilities to engage with various types of data representation. For example, requiring students to interpret a complex graph and draw conclusions based on the data not only strengthens their problem-solving skills but also reduces the likelihood of receiving accurate assistance from an LLM.

This approach to exam design aligns with broader concerns regarding the vulnerability of neural network classifiers to adversarial attacks. Athalye et al. (2018) demonstrated the existence of robust adversarial examples, which are adversarial inputs that remain adversarial over a chosen distribution of transformations [2]. They developed an algorithm for synthesizing these examples and applied it to create three-dimensional (3D) adversarial objects, which were printed using low-cost, commercially available 3D printing technology. Their results showed that these physical adversarial objects were consistently misclassified by a standard ImageNet classifier over various angles, viewpoints, and lighting conditions.

In a similar vein, Alcorn et al. (2019) revealed that deep neural networks (DNNs) often fail to generalize to out-of-distribution (OoD) inputs, including natural, non-adversarial ones that are common in real-world settings [1]. By harnessing 3D renderers and 3D models, the authors presented a framework for discovering DNN failures in response to rendered images. They found that DNNs incorrectly classified 97% of the pose space for objects that were readily recognized in their canonical poses. Furthermore, the study showed that adversarial poses transferred across models and datasets, indicating the vulnerability of DNNs to OoD inputs.

Su et al. (2019) demonstrated that DNNs are vulnerable to low-dimension attacks, such as one-pixel adversarial perturbations, which can be generated using differential evolution (DE) [14]. Their results indicated that 67.97% of the natural images in the Kaggle CIFAR-10 test dataset and 16.04% of the ImageNet (ILSVRC 2012) test images could be perturbed to at least one target class by modifying just one pixel with 74.03% and 22.91% confidence on average. These findings suggest that DNNs are susceptible to attacks even in extremely limited scenarios.

In light of these findings, incorporating non-textual information in exams serves as a means to counteract the potential assistance provided by LLMs, thus maintaining the integrity of the assessment process. By integrating questions that require students to analyze and interpret various forms of data representation, educators can promote critical thinking and problem-solving skills while also creating exams that are more resistant to LLM intervention [1].

## 4 Initial Evalueation of the Proposed methods

While conducting the evaluation of GPT-4's performance, the present study sought to analyze the efficacy of the model in responding to the questions stated above, disregarding the group work assignment. To this end, the questions were tested in their original form as well as with the application of widely recognized prompt optimization techniques[15][17][16]. It is important to note that the GPT-4 ViT model was excluded from the assessment due to its unavailability to the public at the time of writing this paper. Consequently, the analysis relied on the subscription service "ChatGPT Plus" as an alternative means to gauge GPT-4's performance without visual input.

The decision to utilize ChatGPT Plus was informed by the current limited access to GPT-4 API, which is presently restricted by a waitlist. By employing a service that is accessible to the public, the author aimed to ensure that the evaluation remained relevant and useful to a broader student audience. In this regard, the methodology adopted for this study was designed to reflect the real-world experiences of students who might be reliant on publicly available resources.

To assess the quality of the answers generated by the model, the author employed a pass/fail grading system based on the accuracy and correctness of the provided responses. This approach facilitated a comprehensive evaluation of the model's performance, offering valuable insights into its practical utility for academic purposes.

In the case of the chemistry exam problem, the prompt was tested 10 times using various approaches[13][3] to gauge GPT-4's ability to respond accurately. The model refused to answer in all 10 instances, which prompted the decision to discontinue further testing for this particular problem.

Contrastingly, for both the biology and physics exam problems, GPT-4 failed to produce correct answers in 10 out of 10 attempts. Despite this initial failure, further testing was conducted, as the model did not refuse to respond to these questions. The subsequent testing incorporated the use of CoT prompting, a technique that provided GPT-4 with the necessary intermediate steps leading to the answer[15][17][16]. In these cases, GPT-4 was able to generate accurate responses only when the correct chain of thought was inserted. However, the model alone was unable to construct a chain of thought that led to the correct answer in ten out of ten cases. This finding highlights the limitations of GPT-4's problem-solving capabilities, particularly when intermediate steps are not provided explicitly.

It is worth noting that to input the accurate chain of thought, a student would not only need to know the correct answer but also be able to describe each step leading to the solution. This requirement demonstrates a comprehensive understanding of the subject matter, which, in turn, raises questions about the extent to which GPT-4 can genuinely assist students in grasping complex topics. This observation calls for further research to explore the pedagogical implications of using AI language models, such as GPT-4, in academic settings.

## 5 Conclusion

The performance of Large Language Models (LLMs), such as ChatGPT, on exams poses significant implications for academic integrity. While LLMs can generate sophisticated text and perform well in certain situations, their lack of genuine comprehension may result in superficial or erroneous responses. As LLMs continue to evolve and gain prominence, it is crucial for educators to adapt examination design to maintain assessment integrity and promote essential skills development in students. Traditionally, students have been evaluated based on their ability to write long-form essays; however, given the advancements in LLMs and their potential to assist in producing sophisticated written content, it may be time for educators to reconsider this approach. Moreover, exams that do not incorporate modern tools can become inauthentic, failing to represent how professionals would solve problems in real life. By excluding such tools within exams, educators may inadvertently force students to learn and employ outdated problem-solving methods that do not reflect a modern workflow. It is essential to strike a balance between maintaining academic integrity and ensuring that assessments accurately reflect the skills and tools relevant to contemporary professional settings. This article provides guidelines for creating LLM-resistant exams, including safeguarding through content moderation, misdirecting AI models with deliberate inaccuracies, evaluating real-world scenarios beyond the LLM's knowledge base, developing effective distractor options, and incorporating non-textual information in examinations. Furthermore, it underscores the importance of emphasizing soft skills, such as communication, collaboration, leadership, and critical thinking, which are beyond the scope of LLM capabilities. By implementing these strategies, adapting assessments to include modern tools, and shifting the focus towards students' engagement and comprehension, educators can create exams that are more resistant to LLM intervention, ensuring a fair and accurate assessment of students' knowledge and skills. In doing so, they will not only safeguard the integrity of the academic environment but also contribute to the ongoing efforts to address the challenges and opportunities presented by the increasing prevalence of artificial intelligence in education.

## References

[1] Michael A. Alcorn, Qi Li, Zhitao Gong, Chengfei Wang, Long Mai, Wei-Shinn Ku, and Anh Nguyen. Strike (with) a pose: Neural networks are easily fooled by strange poses of familiar objects, 2019.

[2] Anish Athalye, Logan Engstrom, Andrew Ilyas, and Kevin Kwok. Synthesizing robust adversarial examples, 2018.

[3] Hezekiah J. Branch, Jonathan Rodriguez Cefalu, Jeremy McHugh, Leyla Hujer, Aditya Bahl, Daniel del Castillo Iglesias, Ron Heichman, and Ramesh Darwishi. Evaluating the susceptibility of pre-trained language models via handcrafted adversarial examples, 2022.

[4] Educational Testing Service. Gre sample issue task. https://www.ets.org/pdfs/gre/sample-issue-task.pdf, 2022.

[5] John B. Nici. *AP Art History: 5 Practice Tests + Comprehensive Review + Online Practice*. Barron's Educational Series, Hauppauge, NY, 2020.

[6] OpenAI. Gpt-4 model card, 2023.

[7] OpenAI. Gpt-4 technical report, 2023.

[8] OpenAI. Gpt-4 technical report: Appendix a.1, 2023.

[9] OpenAI. Gpt-4 technical report: Appendix a.2, 2023.

[10] OpenAI. Gpt-4 technical report: Appendix a.3, 2023.

[11] OpenAI. Gpt-4 technical report: Appendix b, 2023.

[12] OpenAI. Gpt-4 technical report: Appendix d, 2023.

[13] Fábio Perez and Ian Ribeiro. Ignore previous prompt: Attack techniques for language models, 2022.

[14] Jiawei Su, Danilo Vasconcellos Vargas, and Kouichi Sakurai. One pixel attack for fooling deep neural networks. *IEEE Transactions on Evolutionary Computation*, 23(5):828–841, oct 2019.

[15] Xuezhi Wang, Jason Wei, Dale Schuurmans, Quoc Le, Ed Chi, Sharan Narang, Aakanksha Chowdhery, and Denny Zhou. Self-consistency improves chain of thought reasoning in language models, 2023.

[16] Jason Wei, Xuezhi Wang, Dale Schuurmans, Maarten Bosma, Brian Ichter, Fei Xia, Ed Chi, Quoc Le, and Denny Zhou. Chain-of-thought prompting elicits reasoning in large language models, 2023.

[17] Xi Ye and Greg Durrett. The unreliability of explanations in few-shot prompting for textual reasoning, 2022.